\documentclass[10pt,twocolumn,letterpaper]{article}

\usepackage{wacv}              

\usepackage{graphicx}
\usepackage{amsmath}
\usepackage{amssymb}
\usepackage{booktabs}
\usepackage{float}

\usepackage[pagebackref,breaklinks,colorlinks]{hyperref}

\usepackage[capitalize]{cleveref}
\crefname{section}{Sec.}{Secs.}
\Crefname{section}{Section}{Sections}
\Crefname{table}{Table}{Tables}
\crefname{table}{Tab.}{Tabs.}

\def\wacvPaperID{16} 
\def\confName{WACV}
\def\confYear{2025}

\begin{document}

\title{HFMF: Hierarchical Fusion Meets Multi-Stream Models for Deepfake Detection}

\author{Anant Mehta \quad
Bryant McArthur \quad
Nagarjuna Kolloju \quad 
Zhengzhong Tu\\
Texas A\&M University \\
{\tt\small \{anant\_mehta,\ mcarthur,\ nkolloju,\ tzz\}@tamu.edu}
}
\maketitle

\begin{abstract}
The rapid progress in deep generative models has led to the creation of incredibly realistic synthetic images that are becoming increasingly difficult to distinguish from real-world data. The widespread use of Variational Models, Diffusion Models, and Generative Adversarial Networks has made it easier to generate convincing fake images and videos, which poses significant challenges for detecting and mitigating the spread of misinformation. As a result, developing effective methods for detecting AI-generated fakes has become a pressing concern.
In our research, we propose HFMF, a comprehensive two-stage deepfake detection framework that leverages both hierarchical cross-modal feature fusion and multi-stream feature extraction to enhance detection performance against imagery produced by state-of-the-art generative AI models. The first component of our approach integrates vision Transformers and convolutional nets through a hierarchical feature fusion mechanism. The second component of our framework combines object-level information and a fine-tuned convolutional net model. We then fuse the outputs from both components via an ensemble deep neural net, enabling robust classification performances. We demonstrate that our architecture achieves superior performance across diverse dataset benchmarks while maintaining calibration and interoperability.
The code is available at \url{https://github.com/taco-group/HFMF}.

\end{abstract}

\section{Introduction}
\label{sec:intro}
The rapid advancement of deep generative models has led to the creation of highly realistic and convincing fake digital media, including applications like image generation~\cite{baldridge2024imagen, rombach2022high}, generative editing~\cite{qi2025spire,tu2022maxim}, video generation~\cite{guo2023animatediff,li20244k4dgen}, audio recordings\cite{evans2024stable,kreuk2022audiogen,zhang2022deepfake}, and text generation~\cite{wu2023brief}. These manipulated media, commonly known as deepfakes, have the potential to deceive, defame, and destabilize individuals, communities, and societies. The proliferation of deepfakes has become a pressing concern, as they can be used to spread misinformation, propaganda, and disinformation, posing a significant threat to the foundation of trust in the digital age and modern-day life.

Despite the growing awareness of the deepfake problem, detecting and mitigating their spread remains a challenging task. Existing deepfake detection techniques often rely on finding specific trends or artifacts produced by generative models, but proficient attackers may easily bypass these methods\cite{chadha2021deepfake}. Furthermore, the rapid growth of generative models has sparked a competition between deepfake detectors and deepfake producers, making it harder and harder to create efficient detection approaches\cite{chadha2021deepfake}.

Recent studies have highlighted the limitations of existing deepfake detection approaches, which often suffer from a generalization gap, failing to detect new, unseen deepfake attacks. Furthermore, the increasing computational complexity of deep learning-based detection models has raised concerns about their deployability in real-world applications, particularly on mobile devices with limited computational resources \cite{frank2020leveraging}. In this context, there is a significant need for a novel, efficient, and robust deepfake detection framework that can bridge the gap between current approaches and the rapidly evolving landscape of deepfakes. 

To address this challenge, we propose a novel deepfake detection framework, which we dub HFMF, that combines hierarchical cross-modal feature fusion and multi-stream feature extraction (See Fig. \ref{fig:ensemble}). The first component of our approach leverages Vision Transformers (ViTs) \cite{dosovitskiy2020image,liu2021swin,tu2022maxvit} and Convolutional Neural Networks (ConvNets) \cite{koonce2021resnet,liu2022convnet} in a hierarchical feature fusion mechanism to effectively capture both \textbf{local artifacts} and \textbf{global inconsistencies} introduced by deepfake manipulations. The second component integrates edge features, object-level context, and general image-based features extracted using a couple of lightweight expert models, including Sobel edge detection, YOLOv8 \cite{reis2023real} object detection, and a fine-tuned Xception \cite{chollet2017xception} model. The outputs from these components are fused through an ensemble deep neural network, enabling robust and accurate classification. Another key aspect of our architecture is its low number of trainable parameters, which is achieved by using various pre-trained models in both modules. Our results demonstrate that increased complexity does not necessarily lead to better performance, and our model achieves competitive accuracy while maintaining high computational efficiency.

\section{Related Work}
\subsection{Visual Generative Models}
Visual generative models, including Generative Adversarial Network (GANs), Variational Autoencoders (VAEs), and diffusion models, have made significant strides in image generation. Latent diffusion models, such as those introduced by Rombach $et$ $al.$ \cite{rombach2022high}, generate high-resolution images by modeling complex data distributions in a latent space, enabling efficient synthesis of realistic images. Diffusion models have also evolved into highly effective tools for generating diverse media; for example, Ho $et$ $al.$ \cite{ho2022video} presented video diffusion models, which extend image diffusion models to temporal data. In the context of personalized generation, Guo $et$ $al.$ \cite{guo2023animatediff} introduced Animatediff, enabling animation with text-to-image diffusion models without extensive tuning. More specialized models, such as Audiogen \cite{kreuk2022audiogen}, target audio generation, leveraging similar principles, while 4k4dgen \cite{li20244k4dgen} focuses on panoramic 4D generation at ultra-high resolutions. These advancements have enhanced synthetic content's diversity and resolution, revolutionizing static and dynamic image generation tasks. 

Moreover, powerful pre-trained visual generative models like Stable Diffusion~\cite{evans2024stable} have facilitated strong generative editing methods that can easily manipulate or edit copyrighted or personal images to spread fake news~\cite{brooks2023instructpix2pix,qi2025spire,zhang2023adding,mei2024codi,zhu2024mwformer}.
Without granting permission from the original publishers, these unauthorized edits, if leveraged by malicious hackers, can cause severe ethical or privacy outcomes~\cite{wang2024edit}.
Since these generative methods are highly robust, there is a pressing need for effective techniques to detect deepfakes.

\subsection{Deepfake Detection Techniques}
\label{sec:LR}
One issue in current deepfake detection models exposed by Cavia $et$ $al.$ is inherent biases in the commonly used datasets. Previous datasets used are produced by a single SOTA generative model and also include the same file format, resolution, compression encoding, and a few known simulated post-processing augmentations. It was shown that the models in part learn these hidden biases and do not generalize well to images found in the wild. To resolve this, Cavia $et$ $al.$ released the WildRF \cite{cavia2024real} dataset which contains images found in the wild on three social media platforms: Reddit, X (formerly Twitter), and Facebook. 

Recent advancements in deepfake detection have primarily focused on addressing generative models like VAEs and GANs. For example, Lanzani $et$ $al.$ \cite{Lanzino_2024_CVPR} proposed a lightweight Binary Neural Network (BNN) approach that augments RGB images with FFT and LBP features, achieving real-time detection with minimal accuracy loss and efficient memory usage. Similarly, Coccomini $et$ $al.$ \cite{coccomini2022combining} introduced Transformer-based architectures combining EfficientNet and cross-attention mechanisms, which outperformed traditional networks with lower computational demands. Other techniques, such as Zhao $et$ $al.$'s \cite{zhao2021multi} multi-attentional network with attention-guided augmentation and Bonettini $et$ $al.$'s \cite{bonettini2021video} siamese training with EfficientNetB4, demonstrated state-of-the-art results on popular datasets, although they often relied heavily on dataset-specific training.

As diffusion models become more prominent in generating highly realistic deepfakes, detection methods face increasing challenges. Techniques like Animollah $et$ $al.$'s \cite{khormali2023selfsupervisedgraphtransformerdeepfake} graph-based framework and vision Transformers have shown promising results with high accuracy across diverse datasets. Additionally, Shiohara $et$ $al.$ \cite{shiohara2022detectingdeepfakesselfblendedimages} developed a method using Self-Blended Images (SBI) and EfficientNet for detecting statistical inconsistencies, achieving competitive cross-dataset performance. However, existing works often struggle to generalize across datasets and adapt to rapidly evolving generative technologies, leaving a critical bottleneck in handling deepfakes created with the latest diffusion-based models.

We will fuse together several key components and ideas of these works to develop a novel ensemble for accurate deepfake detection.

\begin{figure}[t]\captionsetup[subfigure]{font=footnotesize}
\begin{center}
\includegraphics[width=8.2cm, height=8.5cm]{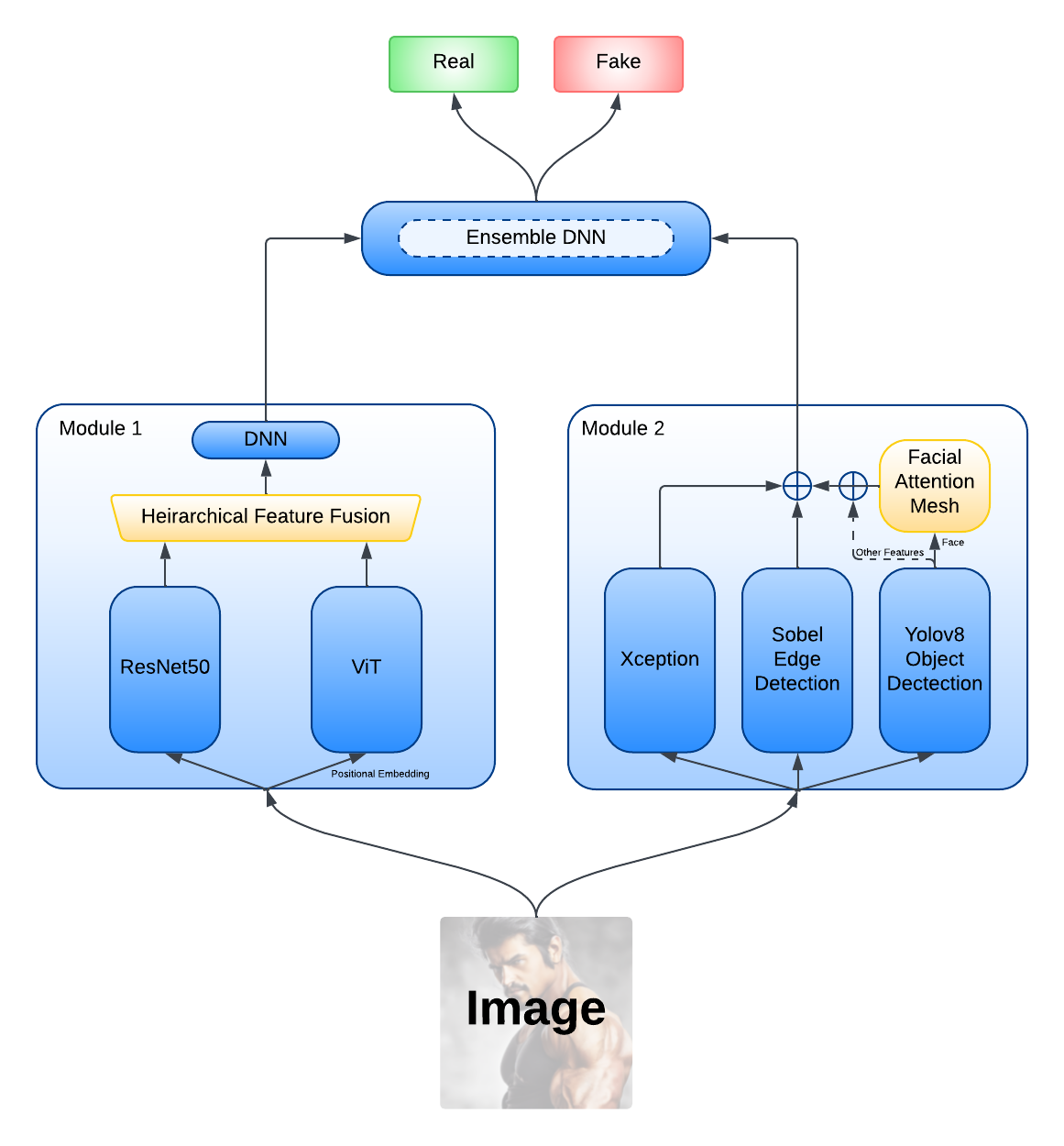}
\caption{Our proposed dual stage ensemble network HFMF.}
\label{fig:ensemble}
\vspace{-6mm}
\end{center}
\end{figure}

\section{Proposed Method}
In this section, we present the proposed deepfake detection framework, HFMF, which consists of two complementary modules: a hierarchical cross-modal feature fusion mechanism and a multi-stream feature extraction pipeline. To enhance the reliability of the first module, we apply a calibration technique to address the known issue of miscalibration in deep ResNets, ensuring that uncertainty does not compromise detection performance. Simultaneously, we have designed the second module to be explainable, enabling the interpretability of the model's predictions. Additionally, we generate explainability maps at the end of this module to provide insights into the regions of the input image that contributed most to the final classification, further validating the robustness and transparency of the proposed framework. Figure \ref{fig:ensemble} shows the final model.
\subsection{Module 1: Hierarchical Feature Fusion}
\paragraph{Semantics-aware feature extraction}
Given an input image \( X \in \mathbb{R}^{H \times W \times 3} \), the ViT model, specifically \( \text{ViT}_{\text{Base-16}} \)\cite{dosovitskiy2020image}, pre-trained on ImageNet \cite{deng2009imagenet}, partitions \( X \) into \( N \) patches of size \( p \times p \times 3 \). Each patch is linearly embedded to form a 768-dimensional feature vector, yielding:
\begin{equation}
    E_{\text{ViT}} = \text{ViT}_{\text{fine-tuned}}(X), \quad E_{\text{ViT}} = [e_1, e_2, \ldots, e_N]
\end{equation}
where \( e_i \in \mathbb{R}^{768} \).
Enjoying the global awareness of the attention mechanisms, the ViT extracted features to capture the contextual or semantic information of the input image, which will be leveraged to fuse with other feature representations.

\begin{figure}[H]\captionsetup[subfigure]{font=footnotesize}
\begin{center}
\includegraphics[width=8.5cm, height=9cm]{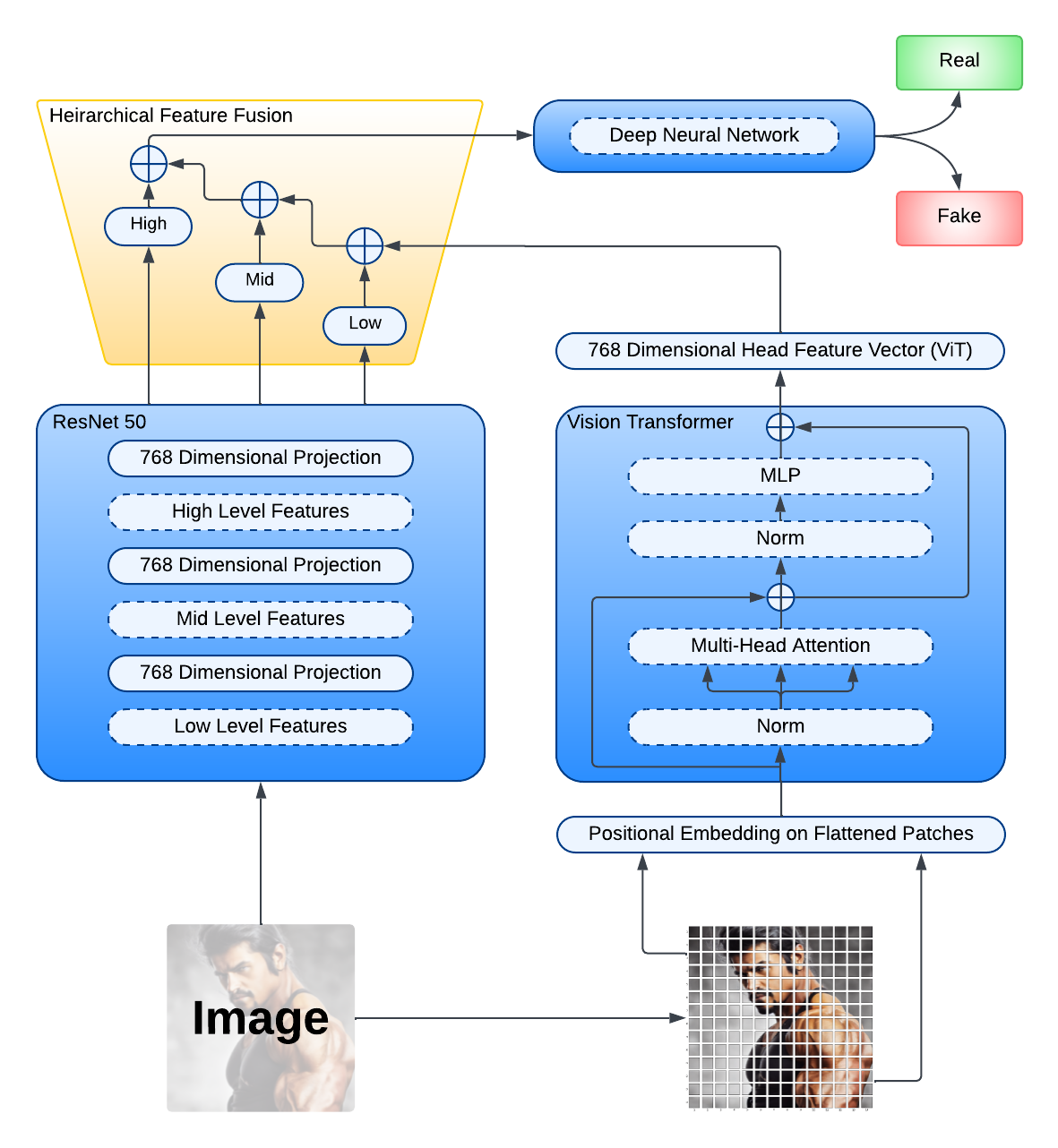}
\caption{Module 1: Hierarchical Cross Feature Fusion.}
\label{fig:m1}
\end{center}
\vspace{-5mm}
\end{figure}

\paragraph{Multi-scale feature extraction}
DeepFake imagery often exhibits abnormalities in local regions, e.g., the famous ``finger issues'' in many diffusion models \cite{kingma2021variational}.
Thus, to account for these local artifacts, we employ the pre-trained ResNet to extract intermediate feature representations from multiple stages, each capturing a different abstraction level:
\begin{align}
    F_{\text{1}},F_{\text{2}},F_{\text{3}} &= \text{ResNet}(X),
\end{align}
Here, \( F_{\text{1}}, F_{\text{2}}, F_{\text{3}} \) represent feature maps at different spatial resolutions, respectively, with \( F_{\text{1}} \) capturing low-level features and \( F_{\text{3}} \) capturing relatively higher-level features.
As DeepFake-type distortions may exhibit at a different scale, according to the generative capabilities of the models, we exploit all the ConvNet features to conduct the final detection.

\paragraph{Hierarchical Attention-based Fusion}
We combine the global contextual features from the ViT branch and the local hierarchical features from the ResNet branch using a hierarchical dot-product similarity mechanism. More specifically, for each feature map \( F_{\text{i}} \), we first reshape it to be in the same resolution (\( \mathbb{R}^{(H \cdot W) \times 768} \)). Then we compute the dot-product similarity using an attention model:
\begin{equation}
    S_{\text{lvl}} = \frac{Q K^{\top}}{\sqrt{d_k}}, \quad d_k = 768,
\end{equation}
where initially \( Q = E_{\text{ViT}} \in \mathbb{R}^{N \times 768} \), \( K = V = F_{\text{i}} \in \mathbb{R}^{(H \cdot W) \times 768} \), and \( S_{\text{lvl}} \in \mathbb{R}^{N \times (H \cdot W)} \).
The hierarchical dot-product similarity operation (HDS) is performed as follows:
\begin{align}
    \text{HDS}(Q, K) &= \text{softmax}(S_{\text{lvl}}) V, \\
    Z_{\text{low}} &= \text{HDS}(E_{\text{ViT}}, F_{\text{low}}) \in \mathbb{R}^{N \times 768}, \\
    Z_{\text{mid}} &= \text{HDS}(Z_{\text{low}}, F_{\text{mid}}) \in \mathbb{R}^{N \times 768}, \\
    Z_{\text{high}} &= \text{HDS}(Z_{\text{mid}}, F_{\text{high}}) \in \mathbb{R}^{N \times 768}.
\end{align}

The final representation, \( V_{\text{final}} = Z_{\text{high}} \), integrates multi-scale information fused incrementally with the global context-aware features, capturing a strong representation power by using two types of neural nets. The fused vector \( V_{\text{final}} \in \mathbb{R}^{N \times 768} \) serves as the input to the deep neural network for further processing and obtaining raw logits. Figure \ref{fig:m1} depicts this module. 

\paragraph{Uncertainty Calibration}
In deep learning models, predicted probabilities often suffer from miscalibration, where the model's confidence does not reflect the true likelihood of correctness. To address this, we apply \textit{Platt Scaling} \cite{phelps2024using}, which adjusts the probability estimates using a logistic regression model on the raw logits \( z \) from the network. The calibrated probability \( \hat{p} \) is given by:

\begin{equation}
    \hat{p} = \frac{1}{1 + \exp-(A z + B)},
\end{equation}
where \( A \) and \( B \) are learned parameters that minimize the negative log-likelihood. We use expected calibration error (ECE) to measure miscalibration.
Calibration is crucial after the hierarchical fusion in Module 1, as deep residual networks often produce overconfident predictions \cite{phelps2024using}.
\begin{equation}
    \text{ECE} = \frac{1}{N} \sum_{i=1}^{N} \frac{|B_i|}{N} \left| \text{acc}(B_i) - \text{conf}(B_i) \right|,
\end{equation}

\noindent where:
\begin{itemize}
    \item \( |B_i| \) is the number of samples in the \( i \)-th bin,
    \item \( N \) is the total number of samples,
    \item \( \text{acc}(B_i) \) is the accuracy in the \( i \)-th bin,
    \item \( \text{conf}(B_i) \) is the average confidence in the \( i \)-th bin.
\end{itemize}

\subsection{Module 2: Multi-Stream Local-Feature Extraction }
\begin{figure}[t]\captionsetup[subfigure]{font=footnotesize}
\begin{center}
\includegraphics[width=8cm, height=10cm]{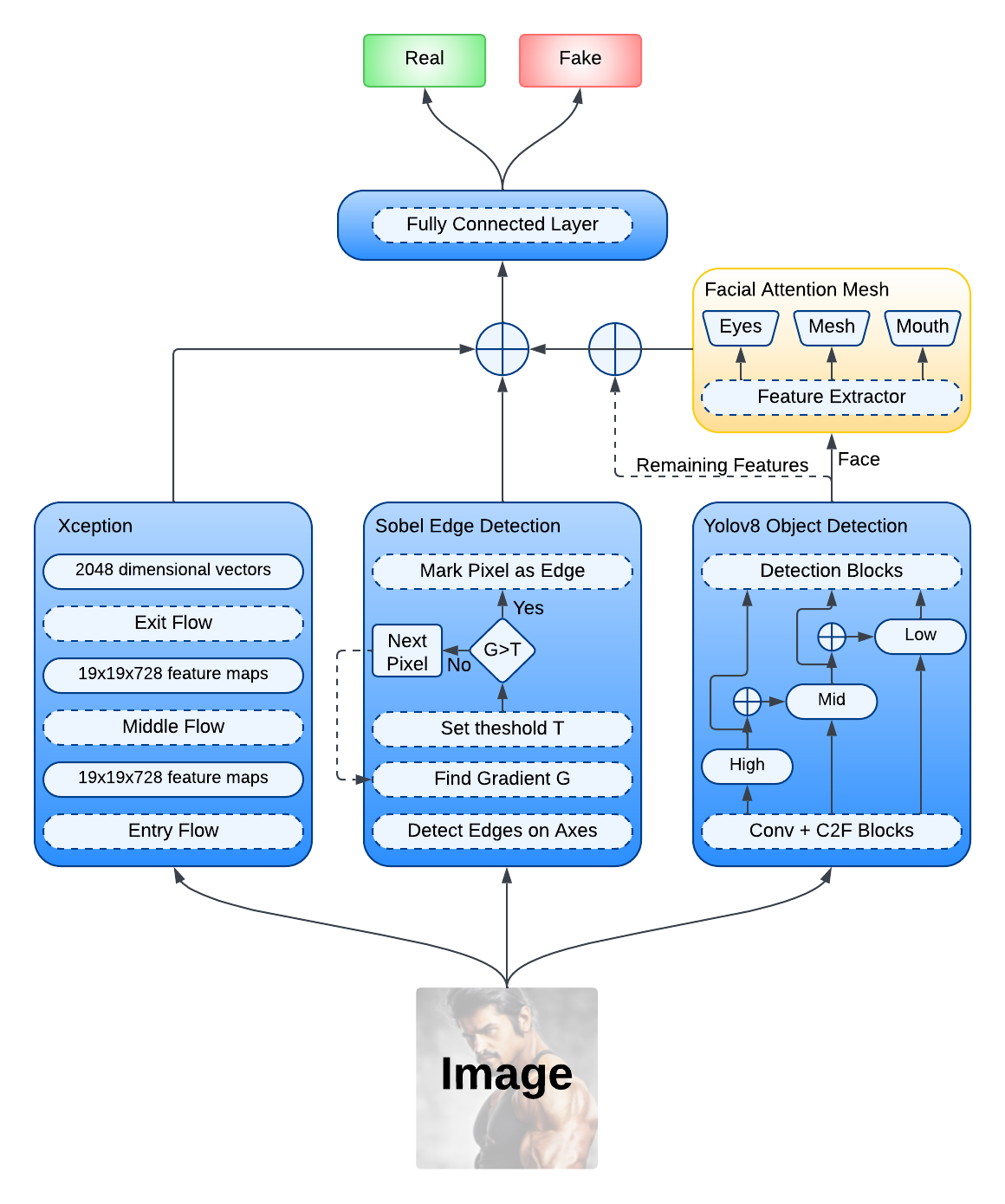}
\caption{Module 2: Multi-Stream Local Feature Extraction.}
\label{fig:m2}
\end{center}
\vspace{-5mm}
\end{figure}

DeepFake media also shows artifacts on small, localized regions, including edges, faces, and small objects, to name a few \cite{chai2020makes}.
Here, we propose a three-branched framework as described in figure \ref{fig:m2}, to capture a variety of localized features using lightweight specialized expert models.

\paragraph{Object and Facial Region Extraction}  
We employ the YOLOv8 \cite{reis2023real} to process the full image \( X \). If a face is detected, cropped regions \( X_f \) are extracted and layered with a Facial Attention Mesh using MediaPipe \cite{lugaresi2019mediapipe} to capture facial features like the mouth and eyes. These features are concatenated with object features \( X_{\text{objects}} \), forming the final representation:
\[
F_X^{\text{YOLO}} = X_f + X_{\text{objects}}.
\]
\( X_{\text{objects}} \) refers to rest of the features from the image, excluding the face.
This module not only emphasizes facial features but also addresses background perturbations to improve robustness in feature representation.

\paragraph{Texture Artifact Embedding}  
The Sobel filter \cite{996} operates on \( X \) to compute gradients in horizontal (\( G_x \)) and vertical (\( G_y \)) directions:
\[
G = \sqrt{G_x^2 + G_y^2}.
\]
These gradients enhance edge detection, producing edge-enhanced images \( X_s \), which emphasize subtle texture-based manipulation artifacts. The features extracted here, \( F_X^{\text{Sobel}} \), are passed to the final decision-making layer to ensure texture-related cues are captured.

\paragraph{Fine-Grained Feature Extraction}  
We utilize the XceptionNet \cite{chollet2017xception}, pre-trained on the ImageNet dataset, processes \( X \) to extract embeddings \( F_X^{\text{Xception}} \in \mathbb{R}^{d_X} \). These embeddings encode fine-grained manipulation patterns and global contextual cues. The module ensures that even nuanced facial alterations or deepfake features are effectively learned.

\subsubsection{Integration Across Modules}  
The outputs from all three modules—\( F_X^{\text{YOLO}} \), \( F_X^{\text{Sobel}} \), and \( F_X^{\text{Xception}} \)—are concatenated and fed into a fully connected layer for final classification:
\[
F_X = F_X^{\text{YOLO}} \oplus F_X^{\text{Sobel}} \oplus F_X^{\text{Xception}},
\]
where \( \oplus \) denotes the concatenation operation. This integration ensures a comprehensive feature representation, capturing spatial, texture, and contextual features for distinguishing real versus fake inputs.

\subsection{Ensemble Decision Making}
HFMF, our ensemble model, depicted in Fig.~\ref{fig:ensemble}, integrates predictions from two specialized modules to enhance overall performance and robustness, using logits of the final neural networks in both cases. Module one combines global features from Vision Transformer (ViT) and local features from ResNet50 through a hierarchical dot-product similarity mechanism. The resulting multi-scale representation \( V_{\text{final}} \) is calibrated using Platt Scaling to correct overconfident predictions, producing reliable logits for a prediction. Module two extracts multi-stream features using YOLOv8 for facial regions, Sobel filtering for texture artifacts, and XceptionNet for fine-grained manipulation patterns. The fused feature vector \( F_{\text{X}} \) from Module two is processed through a neural network to generate logits. The respective output logits from these modules are fused through a weighted aggregation mechanism using a deep neural network, leveraging complementary strengths of different feature representations. 

This ensemble strategy ensures improved accuracy and interpretability, as each module contributes domain-specific insights while mitigating individual weaknesses.

\section{Experiments}

\begin{table*}[h]
\centering
\setlength{\tabcolsep}{13pt}
\begin{tabular}{lccccc}
\toprule
\textbf{Method} & \textbf{Facebook} & \textbf{Reddit} & \textbf{Twitter} & \textbf{Mean} & \textbf{Validation} \\
\midrule
LaDeDa \cite{cavia2024real} & 81.9\% & 91.8\% & 83.3\% & 85.7\% & - \\
Tiny-LaDeDa \cite{cavia2024real} & 80.7\% & 84.5\% & 82.3\% & 82.5\% & - \\
CNNDet \cite{wang2020cnn} & 70.6\% & 75.4\% & 71.4\% & 72.5\% & - \\
PatchFor \cite{chai2020makes} & 77.1\% & 87.8\% & 81.6\% & 82.2\% & - \\
CLIP \cite{ojha2023towards} & 78.4\% & 80.8\% & 78.1\% & 79.1\% & - \\
NPR \cite{tan2024rethinking} & 76.6\% & 89.8\% & 79.5\% & 81.9\% & - \\
\midrule
\textbf{Module 1 (M1)} & 86.6\% & \textbf{93.0\%} & \textbf{86.0\%} & 88.5\% & 91.7\% \\
\textbf{Module 2 (M2)} & 81.9\% & 90.9\% & 85.4\% & 86.0 \% & 88.4\% \\
\textbf{HFMF (M1+M2)} & \textbf{86.9\%} & 92.3\% & 85.8\% & \textbf{89.4 \%} & \textbf{92.7\%} \\
\bottomrule
\end{tabular}
\caption{Accuracy comparison of different methods. All methods are trained on the WildRF training set and evaluated on the three subsets of the WildRF test set. Our results show remarkable improvement across all test sets over previous methods.}
\label{tab:accuracy_comparison}
\end{table*}


\begin{table*}[h]
\centering
\setlength{\tabcolsep}{10pt}
\begin{tabular}{lcccc}
\toprule
\textbf{Dataset} & \textbf{ECE (Uncalibrated)} & \textbf{ECE (Calibrated)} & \textbf{Percentage Decrease ($\downarrow$)} \\
\midrule

WildRF (Train) & 0.0580 & 0.0473 & 18.97\% \\
WildRF (Val)   & 0.1017 & 0.0921 & 9.48\%  \\
\midrule
WildRF (FB)    & 0.1549 & 0.1330 & \textbf{14.13\%} \\
WildRF (Reddit) & 0.0830 & 0.0666 & \textbf{19.79\%} \\
WildRF (Twitter) & 0.1668 & 0.1399 & \textbf{16.14\%} \\
\bottomrule
\end{tabular}
\caption{ECE on Datasets after Calibration of Module 1.}
\label{tab:calibration_results}
\end{table*}

\subsection{Experimental Setup}
We utilized NVIDIA RTX 6000 GPU nodes for training and evaluation of our three models, Module 1, Module 2 and the final HFMF ensemble. We trained final ensemble with early stopping to prevent overfitting. This setup was optimized to balance computational efficiency and model performance. Additionally, we employed a two-stream architecture, where Module One and Module Two were trained separately before ensembling.

\subsection{Datasets and Benchmarks}
One key to a strong deepfake detection model is training with data from state-of-the-art Generative AI models. Even the best architecture cannot compensate for outdated training data in identifying realistic, high-quality fakes.

For this reason, we use WildRF dataset released by Cavia $et$ $al.$ \cite{cavia2024real} as our primary benchmark. The WildRF dataset is compiled of images, real and fake, found on social media websites Facebook (FB), Reddit (R), and X (formerly Twitter) in 2024 (the year our experiments were conducted). The AI generated images reflect the high quality of the most SOTA models available today including Open AI's DALL-E 3 \cite{betker2023improving}, Google DeepMind's Imagen3 \cite{baldridge2024imagen}, X AI's Grok-2 \cite{xai_grok2} and more, as well as finetuned versions of these generative models on specialized datasets, making it nearly imperceptible in some cases for the human eye to recognize the deepfakes. Furthermore, WildRF inherently eliminates several biases, such as JPEG compression, found in other generated image datasets. Also, this dataset includes images generated from unknown synthesizer models, making it more practical for real-world scenarios.


A main concern for deepfakes is identity and misinformation attacks with face generation and manipulation. To address this concern, the second benchmark we use is CollabDif \cite{huang2023collaborative}. CollabDif uses a recent multi-modal approach fusing Latent Diffusion Models (LDMs) and Variational AutoEncoders (VAEs) for generating face images as proposed by Huang et al. in 2023. The model takes both textual prompts and facial segmentations for more control in generating and editing photo realistic images.

Together, the WildRF and CollabDiff datasets ensure comprehensive testing of our approach. WildRF covers both facial and non-facial manipulations, while CollabDiff focuses on high-quality facial deepfake generation and editing.


\subsection{Main Results}
\begin{figure}[H]
\centering
\includegraphics[width=\linewidth, keepaspectratio]{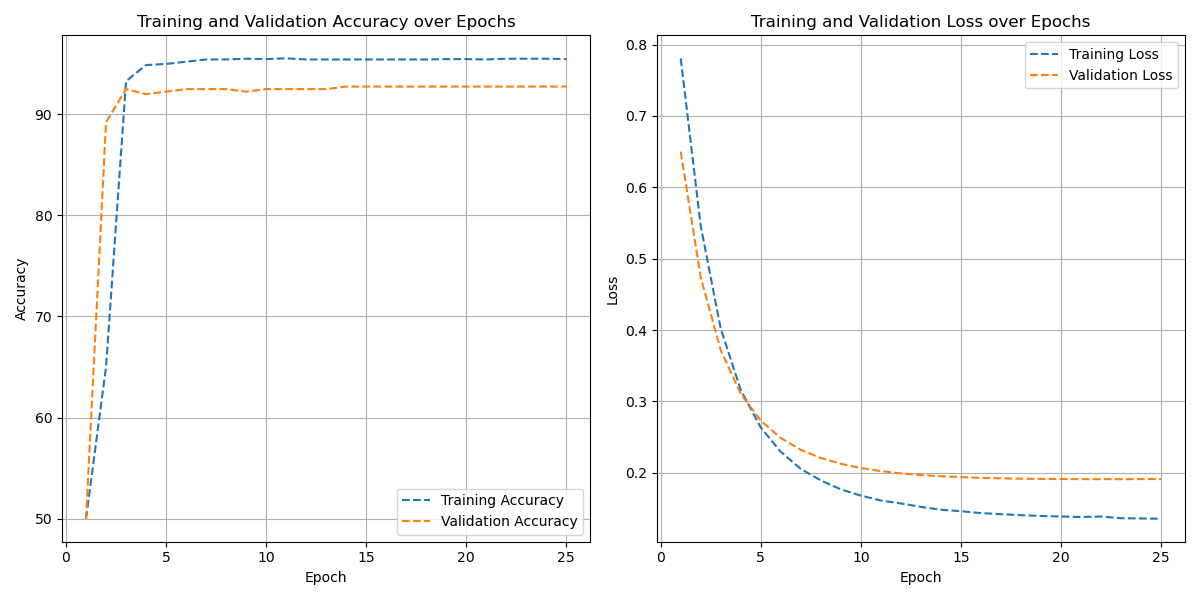}
\caption{Training and Validation curve for HFMF (WildRF).}
\label{fig:graph}
\end{figure}

We conducted extensive experiments on the WildRF dataset, evaluating the model on three distinct test subsets: Facebook, Reddit, and Twitter, as well as the validation set. Additionally, we tested on the CollabDiff validation set, which is generated using latent diffusion models. Despite its small size, CollabDiff consistently yielded nearly perfect results, a trend corroborated by prior works in the literature.

Our experiments evaluated the performance of Module One, Module Two, and the final ensemble. In Module One, a Vision Transformer (ViT) architecture was utilized with a patch size of $16 \times 16$, effectively capturing hierarchical visual features. Module Two employed a multi-stream model incorporating Sobel edge detection and XceptionNet, enhancing spatial and contextual feature fusion. These modules were independently trained before combining their outputs in the final ensemble. 

For the final ensemble (HFMF), we set 100 epochs for both the datasets, employing early stopping. For example, in the case of WildRF the training halted after 25 epochs. The training process achieved a Train Accuracy of 95.43\%, a Validation Loss of 0.1911, and a Validation Accuracy of 92.7\%. Figure \ref{fig:graph} illustrates the training and validation accuracy and loss curves for the HFMF architecture.

Table \ref{tab:accuracy_comparison} presents the accuracy comparisons for various methods tested on the WildRF test subsets. Module One achieved accuracy rates of 86.6\% on Facebook, 93.0\% on Reddit, and 86.0\% on Twitter, with a mean accuracy of 88.5\% and a validation accuracy of 91.7\%. Module Two achieved accuracy rates of 81.9\% on Facebook, 90.9\% on Reddit, and 85.4\% on Twitter, with a mean accuracy of 86.0\% and a validation accuracy of 88.4\%. The final ensemble which we call HFMF further enhanced these results, achieving 86.9\% on Facebook, 92.3\% on Reddit, and 85.8\% on Twitter, with a mean accuracy of 89.4\% and a validation accuracy of 92.7\%. 
\begin{table}[h]
\centering
\setlength{\tabcolsep}{6pt}
\begin{tabular}{lcccc}
\toprule
\textbf{Dataset} & \textbf{Acc.} & \textbf{Prec.} & \textbf{Rec.} & \textbf{F1} \\
\midrule
WildRF (FB) \cite{cavia2024real} & {86.9\%} & 0.92 & 0.81 & 0.86 \\
WildRF (R) \cite{cavia2024real} & 92.3\% & 0.95 & 0.91 & 0.93 \\
WildRF (X) \cite{cavia2024real} & 85.8\% & 0.97 & 0.81 & 0.88 \\
\midrule
CollabDif (VAL) \cite{huang2023collaborative} & {100\%} & {1.00} & {1.00} & {1.00} \\
\bottomrule
\end{tabular}
\caption{Classification scores across different test sets using our final HFMF (M1+M2).}
\label{tab:metric_comparison}
\end{table}
\begin{table}[h]
\centering
\setlength{\tabcolsep}{10pt}
\begin{tabular}{lccccc}
\toprule

\textbf{Method} & \textbf{CollabDif(VAL)} \\
\midrule

\textbf{Module 1 (M1)} & 100\% \\
\textbf{Module 2 (M2)} & 99.50\% \\
\midrule

\textbf{HFMF (M1+M2)} & \textbf{100\%} \\
\bottomrule
\end{tabular}
\caption{Proposed methods' results on CollabDif.}
\label{tab:accuracy_comparison_cd}
\end{table}
Table \ref{tab:metric_comparison} highlights the classification metrics, including accuracy (Acc.), precision (Prec.), recall (Rec.), and F1 scores, across different test subsets. On the CollabDif validation set, as shown in Table \ref{tab:accuracy_comparison_cd}, Module One and the final ensemble achieved 100\% accuracy, while Module two closely followed at 99.50\%. These results emphasize the effectiveness of hierarchical fusion and multi-stream modeling in HFMF in delivering robust and accurate deepfake detection. In the following sections, we also discuss the ablation study results.

\begin{table*}[h]
\centering
\setlength{\tabcolsep}{4pt}
\begin{tabular}{lcccccc}
\toprule

\textbf{Method} & \textbf{ViT-B/16} & \textbf{ResNet50} & \textbf{Facebook} & \textbf{Reddit} & \textbf{Twitter} & \textbf{Mean} \\
\midrule
\textbf{Pre-trained ViT-B/16} & \checkmark &$\times$  & 78.1\% & 89.4\% & 85.0\% & 84.1\% \\
\textbf{Pre-trained ResNet50} & $\times$  & \checkmark & 84.4\% & \textbf{94.6}\% & 84.0\% & 87.7\% \\
\textbf{Module 1 (ViT-B/16+ResNet50)} & \checkmark & \checkmark & 86.6\% & 93.0\% & \textbf{86.0\%} & 88.5\% \\
\midrule

\textbf{HFMF (Module 1+Module 2)} & \checkmark & \checkmark & \textbf{86.9\%} & 92.3\% & 85.8\% & \textbf{89.4\%} \\
\bottomrule
\end{tabular}
\caption{Ablation studies for Module 1 (ViT-B/16: Vision Transformer, ResNet50) on WildRF.}
\label{tab:module1}
\end{table*}

\begin{table*}[h]
\centering
\setlength{\tabcolsep}{4pt}
\begin{tabular}{lccccccc}
\toprule
\textbf{Method} & \textbf{XCN} & \textbf{Yolo} & \textbf{SOBEL} & \textbf{Facebook} & \textbf{Reddit} & \textbf{Twitter} & \textbf{Mean} \\
\midrule

\textbf{Pre-trained XCN + SOBEL} & \checkmark & $\times$ & \checkmark & 88.2\% & 89.4\% & 78.2\% & 85.3\% \\
\textbf{Yolo + SOBEL} & $\times$ & \checkmark & \checkmark & 66.9\% & 76.8\% & 70.6\% & 71.4\% \\
\textbf{Pre-trained XCN + Yolo} & \checkmark & \checkmark & $\times$ & 80.9\% & 91.6\% & 84.2\% & 85.6\% \\
\textbf{Module 2 (Yolo+XCN+SOBEL)} & \checkmark & \checkmark & \checkmark & 81.9\% & 90.9\% & 85.4\% & 86.0\% \\
\midrule
\textbf{HFMF (Module 1+Module 2)} & \checkmark & \checkmark & \checkmark & \textbf{86.9\%} & \textbf{92.3\%} & \textbf{85.8\%} & \textbf{89.4\%} \\
\bottomrule
\end{tabular}
\caption{Ablation studies for Module 2 (XCN: XceptionNet, Yolo: Yolov8, SOBEL: Sobel Edge Detection) on WildRF.}
\label{tab:module2}
\end{table*}

\subsection{Calibration Results}

For calibrating Module 1 on the datasets, we utilized 500 bins for probability calibration. This approach was driven due to the highly skewed distribution of predicted probabilities, which tended to concentrate around the extreme values close to 0 or 1. Using a large number of bins allowed for finer granularity in capturing and adjusting these probability estimates, thereby improving the model's calibration and ensuring that the predicted probabilities better aligned with the observed outcomes. This approach was particularly critical given the dataset's characteristics and the need for precise probability estimates in downstream tasks. After calibration on the training dataset, the Expected Calibration Error (ECE) was significantly reduced, demonstrating the effectiveness of the calibration method. The uncalibrated and calibrated ECE values for various datasets (training, validation, and social media subsets) are provided below in table \ref{tab:calibration_results}. Relative change or the decrease in ECE is also analyzed.

The table above shows the ECE values for both uncalibrated and calibrated models across various datasets. As seen, calibration significantly reduces the ECE, with the largest decrease observed on the Reddit test dataset (19.79\%) and the smallest on the validation set (9.48\%). These improvements demonstrate the effectiveness of calibration in enhancing model performance, especially when evaluated on real-world data from platforms like Facebook, Reddit, and Twitter present in WildRF dataset.

In the case of the CollabDif dataset, the model's initial calibration yielded an ECE almost 0, indicating perfect calibration, i.e. model's predictions were already well-aligned with true probabilities, and further calibration was unnecessary.
\subsection{Grad-CAM for Feature Visualization}  
\begin{figure}[H]\captionsetup[subfigure]{font=footnotesize}
\begin{center}
\includegraphics[width=8cm, height=5.5cm]{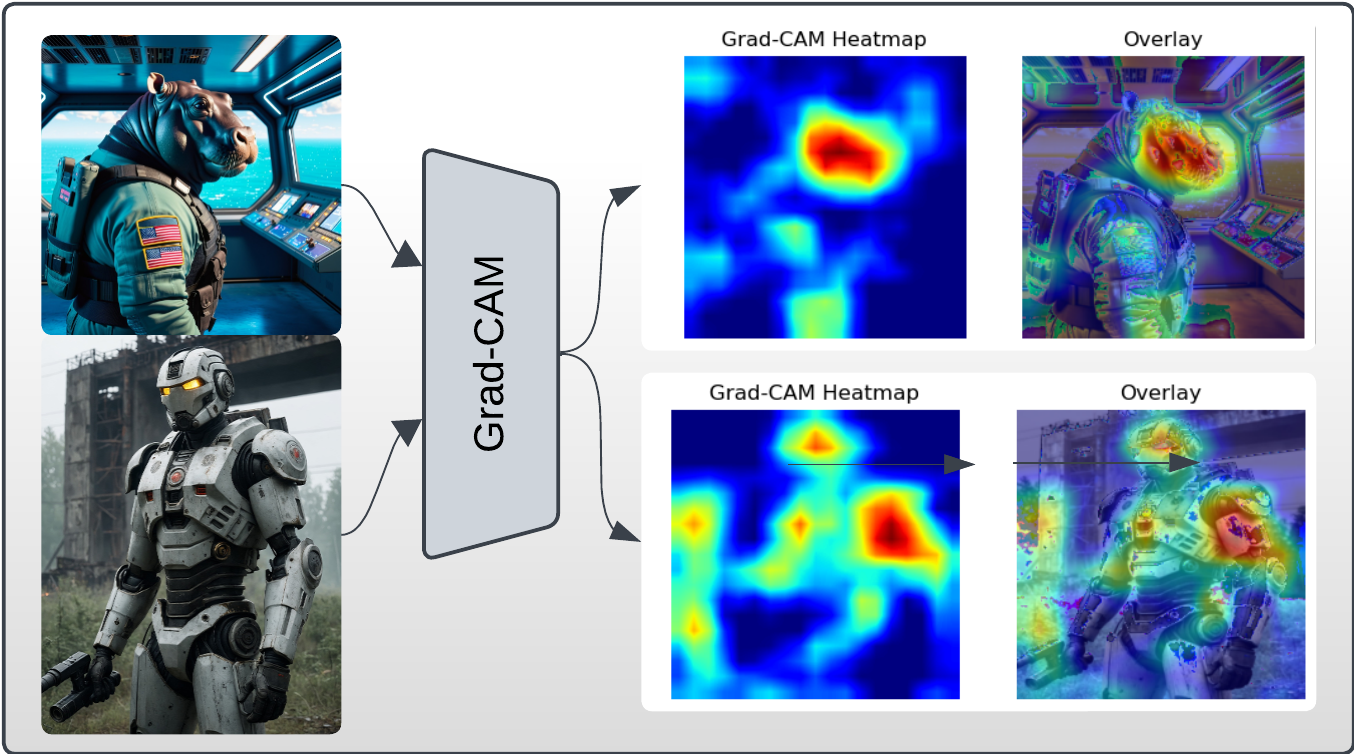}
\caption{Grad-CAM visualization for WildRF}
\label{fig:gradcam}
\end{center}
\end{figure}
\begin{figure}[h]\captionsetup[subfigure]{font=footnotesize}
\begin{center}
\includegraphics[width=7cm, height=7.5cm]{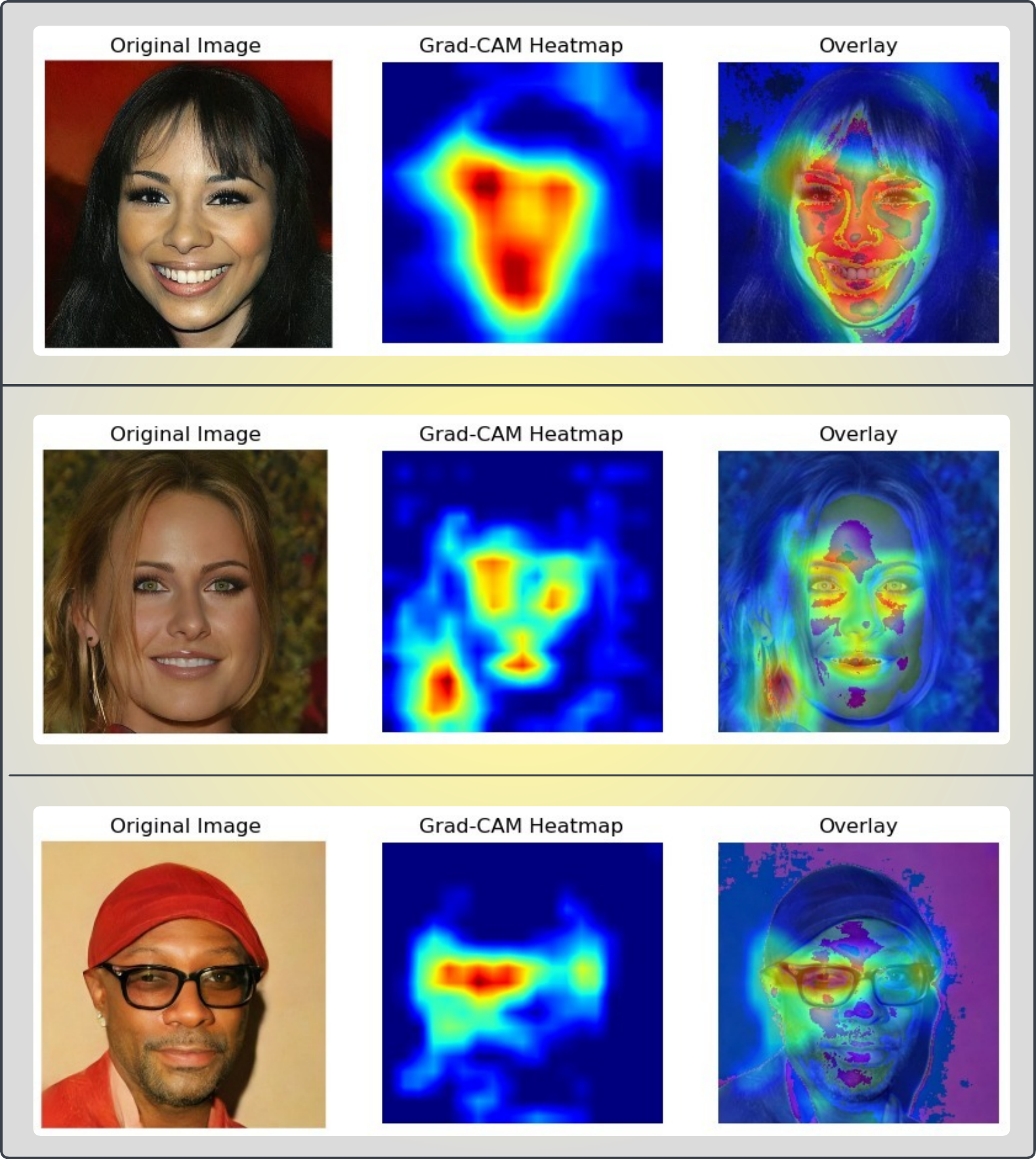}
\caption{Grad-CAM visualization for CollabDif}
\label{fig:gradcam-cd}
\end{center}
\end{figure}

To explain feature extraction in Module 2, we employ Grad-CAM (Gradient-weighted Class Activation Mapping) \cite{Selvaraju_2019}, which generates a localization map \( L_c^{\text{Grad-CAM}} \in \mathbb{R}^{u \times v} \) highlighting regions in the input image \( X \) relevant to the model's decision. For a target class \( c \), Grad-CAM computes the gradients of the class score \( y_c \) (pre-softmax) with respect to feature map activations \( A_k \), and combines them using importance weights \( \alpha_k^c \):
\[
L_c^{\text{Grad-CAM}} = \text{ReLU} \left( \sum_k \alpha_k^c A_k \right),
\]
where ReLU suppresses irrelevant regions. Grad-CAM enhances interpretability by visualizing class-relevant features, as shown in Fig.~\ref{fig:gradcam} and Fig.~\ref{fig:gradcam-cd}. These heatmaps illustrate the rationale behind why Module 2 classified the images in the respective categories.

\subsection{Ablation Studies}  
As shown in Table \ref{tab:module1}, Module 1 (M1), which combines Vision Transformer (ViT-B/16) and ResNet50, achieves a mean accuracy of 88.5\%, surpassing standalone ViT-B/16 (84.1\%) and ResNet50 (87.7\%). This demonstrates the benefit of leveraging complementary features from both architectures using proposed hierarchical cross feature fusion. 

Similarly, Module 2 (M2) as depicted in Table \ref{tab:module2}, which integrates Yolov8, Sobel edge detection, and XceptionNet, achieves a mean accuracy of 86.0\%, outperforming simpler combinations among these models. Notably, the addition of XceptionNet in M2 significantly enhances performance by providing robust feature extraction capabilities, complementing the edge detection of Sobel and the object localization of Yolov8. This synergy highlights the importance of combining diverse model functionalities to achieve superior results across datasets. The inclusion of all three components in M2 allows it to leverage edge detection and object localization alongside deep feature extraction, resulting in more robust performance across datasets.

However, neither M1 nor M2 alone could consistently achieve top performance across all datasets. While M2 performs slightly lower than M1 in terms of mean accuracy (86.0\% vs. 88.5\%), it adds valuable explainability through its integration of Sobel edge detection and object localization, which enhances interpretability. By fusing M1 and M2 in HFMF, the model achieves a mean accuracy of 89.4\%, successfully combining the strengths of both modules. This fusion allows HFMF to deliver superior results while maintaining both certainty and explainability.
\section{Conclusion \& Future Scope}

In this work, we present a robust multi-module dual stream approach for deepfake detection, combining state-of-the-art techniques in feature extraction, hierarchical fusion, and ensemble decision-making. By leveraging calibrated outputs, Grad-CAM explainability, and a combination of modern architectures such as YOLOv8, Sobel filtering, and XceptionNet, our model demonstrates improved reliability and accuracy. The integration of training data from recent Generative AI models ensures adaptability to real-world scenarios. This comprehensive pipeline not only enhances detection performance but also provides interpretability, making it a step forward in addressing the challenges posed by realistic deepfakes.

In future work, this framework can be extended to multimedia forensics by incorporating multi-modal inputs like video, audio, and textual metadata, enabling detection beyond images. The hierarchical fusion mechanism can also address disinformation and credibility assessment in real-world scenarios, such as social media and forensic investigations. We will also focus on improving robustness and real-time efficiency in constrained environments.

Our fully supervised approach will be extended to semi-supervised or unsupervised settings, broadening applicability to real-world scenarios with unseen synthesizers.

\paragraph{Ethical Statement:}
This work aims to address the ethical challenges posed by deepfakes, focusing on combating their misuse for misinformation, fraud, and privacy violations. Our detection framework is intended for responsible applications that promote transparency and safety. We encourage its ethical use and advocate for continued collaboration to mitigate the broader societal impacts of generative AI.

\bibliographystyle{unsrt} 
\bibliography{main} 

\begin{thebibliography}{10}

\bibitem{baldridge2024imagen}
Jason Baldridge, Jakob Bauer, Mukul Bhutani, Nicole Brichtova, Andrew Bunner, Kelvin Chan, Yichang Chen, Sander Dieleman, Yuqing Du, Zach Eaton-Rosen, et~al.
\newblock Imagen 3.
\newblock {\em arXiv preprint arXiv:2408.07009}, 2024.

\bibitem{rombach2022high}
Robin Rombach, Andreas Blattmann, Dominik Lorenz, Patrick Esser, and Bj{\"o}rn Ommer.
\newblock High-resolution image synthesis with latent diffusion models.
\newblock In {\em Proceedings of the IEEE/CVF conference on computer vision and pattern recognition}, pages 10684--10695, 2022.

\bibitem{qi2025spire}
Chenyang Qi, Zhengzhong Tu, Keren Ye, Mauricio Delbracio, Peyman Milanfar, Qifeng Chen, and Hossein Talebi.
\newblock Spire: Semantic prompt-driven image restoration.
\newblock In {\em European Conference on Computer Vision}, pages 446--464. Springer, 2025.

\bibitem{tu2022maxim}
Zhengzhong Tu, Hossein Talebi, Han Zhang, Feng Yang, Peyman Milanfar, Alan Bovik, and Yinxiao Li.
\newblock Maxim: Multi-axis mlp for image processing.
\newblock In {\em Proceedings of the IEEE/CVF conference on computer vision and pattern recognition}, pages 5769--5780, 2022.

\bibitem{guo2023animatediff}
Yuwei Guo, Ceyuan Yang, Anyi Rao, Zhengyang Liang, Yaohui Wang, Yu~Qiao, Maneesh Agrawala, Dahua Lin, and Bo~Dai.
\newblock Animatediff: Animate your personalized text-to-image diffusion models without specific tuning.
\newblock {\em arXiv preprint arXiv:2307.04725}, 2023.

\bibitem{li20244k4dgen}
Renjie Li, Panwang Pan, Bangbang Yang, Dejia Xu, Shijie Zhou, Xuanyang Zhang, Zeming Li, Achuta Kadambi, Zhangyang Wang, Zhengzhong Tu, et~al.
\newblock 4k4dgen: Panoramic 4d generation at 4k resolution.
\newblock {\em arXiv preprint arXiv:2406.13527}, 2024.

\bibitem{evans2024stable}
Zach Evans, Julian~D Parker, CJ~Carr, Zack Zukowski, Josiah Taylor, and Jordi Pons.
\newblock Stable audio open.
\newblock {\em arXiv preprint arXiv:2407.14358}, 2024.

\bibitem{kreuk2022audiogen}
Felix Kreuk, Gabriel Synnaeve, Adam Polyak, Uriel Singer, Alexandre D{\'e}fossez, Jade Copet, Devi Parikh, Yaniv Taigman, and Yossi Adi.
\newblock Audiogen: Textually guided audio generation.
\newblock {\em arXiv preprint arXiv:2209.15352}, 2022.

\bibitem{zhang2022deepfake}
Tao Zhang.
\newblock Deepfake generation and detection, a survey.
\newblock {\em Multimedia Tools and Applications}, 81(5):6259--6276, 2022.

\bibitem{wu2023brief}
Tianyu Wu, Shizhu He, Jingping Liu, Siqi Sun, Kang Liu, Qing-Long Han, and Yang Tang.
\newblock A brief overview of chatgpt: The history, status quo and potential future development.
\newblock {\em IEEE/CAA Journal of Automatica Sinica}, 10(5):1122--1136, 2023.

\bibitem{chadha2021deepfake}
Anupama Chadha, Vaibhav Kumar, Sonu Kashyap, and Mayank Gupta.
\newblock Deepfake: an overview.
\newblock In {\em Proceedings of second international conference on computing, communications, and cyber-security: IC4S 2020}, pages 557--566. Springer, 2021.

\bibitem{frank2020leveraging}
Joel Frank, Thorsten Eisenhofer, Lea Sch{\"o}nherr, Asja Fischer, Dorothea Kolossa, and Thorsten Holz.
\newblock Leveraging frequency analysis for deep fake image recognition.
\newblock In {\em International conference on machine learning}, pages 3247--3258. PMLR, 2020.

\bibitem{dosovitskiy2020image}
Alexey Dosovitskiy.
\newblock An image is worth 16x16 words: Transformers for image recognition at scale.
\newblock {\em arXiv preprint arXiv:2010.11929}, 2020.

\bibitem{liu2021swin}
Ze~Liu, Yutong Lin, Yue Cao, Han Hu, Yixuan Wei, Zheng Zhang, Stephen Lin, and Baining Guo.
\newblock Swin transformer: Hierarchical vision transformer using shifted windows.
\newblock In {\em Proceedings of the IEEE/CVF international conference on computer vision}, pages 10012--10022, 2021.

\bibitem{tu2022maxvit}
Zhengzhong Tu, Hossein Talebi, Han Zhang, Feng Yang, Peyman Milanfar, Alan Bovik, and Yinxiao Li.
\newblock Maxvit: Multi-axis vision transformer.
\newblock In {\em European conference on computer vision}, pages 459--479. Springer, 2022.

\bibitem{koonce2021resnet}
Brett Koonce and Brett Koonce.
\newblock Resnet 50.
\newblock {\em Convolutional neural networks with swift for tensorflow: image recognition and dataset categorization}, pages 63--72, 2021.

\bibitem{liu2022convnet}
Zhuang Liu, Hanzi Mao, Chao-Yuan Wu, Christoph Feichtenhofer, Trevor Darrell, and Saining Xie.
\newblock A convnet for the 2020s.
\newblock In {\em Proceedings of the IEEE/CVF conference on computer vision and pattern recognition}, pages 11976--11986, 2022.

\bibitem{reis2023real}
Dillon Reis, Jordan Kupec, Jacqueline Hong, and Ahmad Daoudi.
\newblock Real-time flying object detection with yolov8.
\newblock {\em arXiv preprint arXiv:2305.09972}, 2023.

\bibitem{chollet2017xception}
Fran{\c{c}}ois Chollet.
\newblock Xception: Deep learning with depthwise separable convolutions.
\newblock In {\em Proceedings of the IEEE conference on computer vision and pattern recognition}, pages 1251--1258, 2017.

\bibitem{ho2022video}
Jonathan Ho, Tim Salimans, Alexey Gritsenko, William Chan, Mohammad Norouzi, and David~J Fleet.
\newblock Video diffusion models.
\newblock {\em Advances in Neural Information Processing Systems}, 35:8633--8646, 2022.

\bibitem{brooks2023instructpix2pix}
Tim Brooks, Aleksander Holynski, and Alexei~A Efros.
\newblock Instructpix2pix: Learning to follow image editing instructions.
\newblock In {\em Proceedings of the IEEE/CVF Conference on Computer Vision and Pattern Recognition}, pages 18392--18402, 2023.

\bibitem{zhang2023adding}
Lvmin Zhang, Anyi Rao, and Maneesh Agrawala.
\newblock Adding conditional control to text-to-image diffusion models.
\newblock In {\em Proceedings of the IEEE/CVF International Conference on Computer Vision}, pages 3836--3847, 2023.

\bibitem{mei2024codi}
Kangfu Mei, Mauricio Delbracio, Hossein Talebi, Zhengzhong Tu, Vishal~M Patel, and Peyman Milanfar.
\newblock Codi: Conditional diffusion distillation for higher-fidelity and faster image generation.
\newblock In {\em Proceedings of the IEEE/CVF Conference on Computer Vision and Pattern Recognition}, pages 9048--9058, 2024.

\bibitem{zhu2024mwformer}
Ruoxi Zhu, Zhengzhong Tu, Jiaming Liu, Alan~C Bovik, and Yibo Fan.
\newblock Mwformer: Multi-weather image restoration using degradation-aware transformers.
\newblock {\em IEEE Transactions on Image Processing}, 2024.

\bibitem{wang2024edit}
Hanhui Wang, Yihua Zhang, Ruizheng Bai, Yue Zhao, Sijia Liu, and Zhengzhong Tu.
\newblock Edit away and my face will not stay: Personal biometric defense against malicious generative editing.
\newblock {\em arXiv preprint arXiv:2411.16832}, 2024.

\bibitem{cavia2024real}
Bar Cavia, Eliahu Horwitz, Tal Reiss, and Yedid Hoshen.
\newblock Real-time deepfake detection in the real-world.
\newblock {\em {arXiv preprint arXiv:2406.09398}}, 2024.

\bibitem{Lanzino_2024_CVPR}
Romeo Lanzino, Federico Fontana, Anxhelo Diko, Marco~Raoul Marini, and Luigi Cinque.
\newblock Faster than lies: Real-time deepfake detection using binary neural networks.
\newblock In {\em Proceedings of the IEEE/CVF Conference on Computer Vision and Pattern Recognition (CVPR) Workshops}, pages 3771--3780, June 2024.

\bibitem{coccomini2022combining}
Davide~Alessandro Coccomini, Nicola Messina, Claudio Gennaro, and Fabrizio Falchi.
\newblock Combining efficientnet and vision transformers for video deepfake detection.
\newblock In {\em International conference on image analysis and processing}, pages 219--229. Springer, 2022.

\bibitem{zhao2021multi}
Hanqing Zhao, Wenbo Zhou, Dongdong Chen, Tianyi Wei, Weiming Zhang, and Nenghai Yu.
\newblock Multi-attentional deepfake detection.
\newblock In {\em Proceedings of the IEEE/CVF conference on computer vision and pattern recognition}, pages 2185--2194, 2021.

\bibitem{bonettini2021video}
Nicolo Bonettini, Edoardo~Daniele Cannas, Sara Mandelli, Luca Bondi, Paolo Bestagini, and Stefano Tubaro.
\newblock Video face manipulation detection through ensemble of cnns.
\newblock In {\em 2020 25th international conference on pattern recognition (ICPR)}, pages 5012--5019. IEEE, 2021.

\bibitem{khormali2023selfsupervisedgraphtransformerdeepfake}
Aminollah Khormali and Jiann-Shiun Yuan.
\newblock Self-supervised graph transformer for deepfake detection, 2023.

\bibitem{shiohara2022detectingdeepfakesselfblendedimages}
Kaede Shiohara and Toshihiko Yamasaki.
\newblock Detecting deepfakes with self-blended images, 2022.

\bibitem{deng2009imagenet}
Jia Deng, Wei Dong, Richard Socher, Li-Jia Li, Kai Li, and Li~Fei-Fei.
\newblock Imagenet: A large-scale hierarchical image database.
\newblock In {\em 2009 IEEE conference on computer vision and pattern recognition}, pages 248--255. Ieee, 2009.

\bibitem{kingma2021variational}
Diederik Kingma, Tim Salimans, Ben Poole, and Jonathan Ho.
\newblock Variational diffusion models.
\newblock {\em Advances in neural information processing systems}, 34:21696--21707, 2021.

\bibitem{phelps2024using}
Nathan Phelps, Daniel~J Lizotte, and Douglas~G Woolford.
\newblock Using platt's scaling for calibration after undersampling--limitations and how to address them.
\newblock {\em arXiv preprint arXiv:2410.18144}, 2024.

\bibitem{chai2020makes}
Lucy Chai, David Bau, Ser-Nam Lim, and Phillip Isola.
\newblock What makes fake images detectable? understanding properties that generalize.
\newblock In {\em Computer Vision--ECCV 2020: 16th European Conference, Glasgow, UK, August 23--28, 2020, Proceedings, Part XXVI 16}, pages 103--120. Springer, 2020.

\bibitem{lugaresi2019mediapipe}
Camillo Lugaresi, Jiuqiang Tang, Hadon Nash, Chris McClanahan, Esha Uboweja, Michael Hays, Fan Zhang, Chuo-Ling Chang, Ming~Guang Yong, Juhyun Lee, et~al.
\newblock Mediapipe: A framework for building perception pipelines.
\newblock {\em arXiv preprint arXiv:1906.08172}, 2019.

\bibitem{996}
N.~Kanopoulos, N.~Vasanthavada, and R.L. Baker.
\newblock Design of an image edge detection filter using the sobel operator.
\newblock {\em IEEE Journal of Solid-State Circuits}, 23(2):358--367, 1988.

\bibitem{wang2020cnn}
Sheng-Yu Wang, Oliver Wang, Richard Zhang, Andrew Owens, and Alexei~A Efros.
\newblock Cnn-generated images are surprisingly easy to spot... for now.
\newblock In {\em Proceedings of the IEEE/CVF conference on computer vision and pattern recognition}, pages 8695--8704, 2020.

\bibitem{ojha2023towards}
Utkarsh Ojha, Yuheng Li, and Yong~Jae Lee.
\newblock Towards universal fake image detectors that generalize across generative models.
\newblock In {\em Proceedings of the IEEE/CVF Conference on Computer Vision and Pattern Recognition}, pages 24480--24489, 2023.

\bibitem{tan2024rethinking}
Chuangchuang Tan, Yao Zhao, Shikui Wei, Guanghua Gu, Ping Liu, and Yunchao Wei.
\newblock Rethinking the up-sampling operations in cnn-based generative network for generalizable deepfake detection.
\newblock In {\em Proceedings of the IEEE/CVF Conference on Computer Vision and Pattern Recognition}, pages 28130--28139, 2024.

\bibitem{betker2023improving}
James Betker, Gabriel Goh, Li~Jing, Tim Brooks, Jianfeng Wang, Linjie Li, Long Ouyang, Juntang Zhuang, Joyce Lee, Yufei Guo, et~al.
\newblock Improving image generation with better captions.
\newblock {\em Computer Science. https://cdn. openai. com/papers/dall-e-3. pdf}, 2(3):8, 2023.

\bibitem{xai_grok2}
XAI.
\newblock Grok2, 2024.

\bibitem{huang2023collaborative}
Ziqi Huang, Kelvin~CK Chan, Yuming Jiang, and Ziwei Liu.
\newblock Collaborative diffusion for multi-modal face generation and editing.
\newblock In {\em Proceedings of the IEEE/CVF Conference on Computer Vision and Pattern Recognition}, pages 6080--6090, 2023.

\bibitem{Selvaraju_2019}
Ramprasaath~R. Selvaraju, Michael Cogswell, Abhishek Das, Ramakrishna Vedantam, Devi Parikh, and Dhruv Batra.
\newblock Grad-cam: Visual explanations from deep networks via gradient-based localization.
\newblock {\em International Journal of Computer Vision}, 128(2), 2019.

\end{thebibliography}

\end{document}